\documentclass[11pt]{article}

\usepackage{acl}

\usepackage{times}
\usepackage{latexsym}

\usepackage[T1]{fontenc}

\usepackage[utf8]{inputenc}

\usepackage{microtype}

\usepackage{inconsolata}

\usepackage{graphicx}
\usepackage{amsmath,amssymb}  
\usepackage{booktabs}         
\usepackage{multirow}         
\usepackage{algorithm}
\usepackage{algorithmic}      
\usepackage{subcaption}
\usepackage{tabularx}
\usepackage{pifont}
\usepackage{xcolor}   
\usepackage[most]{tcolorbox}
\newcommand{\cmark}{\ding{51}}
\newcommand{\xmark}{\ding{55}}
\newcommand{\oldans}[1]{\textcolor{red!80!black}{\textbf{#1}}} 
\newcommand{\newans}[1]{\textcolor{blue!80!black}{\textbf{#1}}} 
\newcommand{\errortext}[1]{\textcolor{orange!90!black}{\textit{#1}}} 
\title{MIRAGE: A Framework for Multi-hop Reasoning
under Dynamic Knowledge Injection}

\author{Heyang Xu\\\And
  Hanwen Hao\\\And
  Haifeng Wang\\\And Haoyi Zhou\\\And Jianxin Li \\}

\begin{document}
\maketitle
\begin{abstract}
Multi-hop Question Answering (MQA) requires large language models (LLMs) to retrieve and synthesize multiple pieces of evidence into a coherent reasoning path, which is essential for complex tasks. However, prior work on knowledge updating for MQA has largely focused on simple fact edits, overlooking a more challenging and realistic setting---\emph{Knowledge Injection}---in which the model must simultaneously accommodate both conflicting updates and entirely novel information. Under this setting, a hop-level answer is often wrong simply because the model trusts the wrong knowledge source: it clings to outdated parametric memory when the injected knowledge conflicts with it (K2), and it cannot answer when the required fact is entirely new (K3).

To investigate this, we introduce NEXUS, the first benchmark tailored for multi-hop reasoning under Knowledge Injection, which explicitly covers both conflicting (K2) and incremental (K3) knowledge. We further propose MIRAGE, an efficient, training-free framework whose core component, the Correlation-aware Dual-path Reasoner (CDR), resolves such per-hop source-trust errors via P(True)-based confidence routing, complemented by a lightweight template-guided corrector for residual structural deviations. Extensive experiments across four LLM families and two benchmarks demonstrate that MIRAGE consistently improves both final-answer accuracy and hop-level reasoning accuracy, establishing a new state of the art in this challenging setting.
\end{abstract}



\section{Introduction}
\label{sec:Introduction}

LLMs can effectively decompose a complex question into sequential sub-questions for multi-hop reasoning \citep{kojima2022large, chen2024plan}, but they often hallucinate in edited-scenario MQA because their parametric knowledge is outdated \citep{patil2025advancing, su2024dragin}---a risk that is especially dangerous in safety-critical domains such as medical diagnosis.

To keep LLMs' knowledge current, memory-based editing, which leverages retrieval-augmented generation (RAG), has emerged as a promising direction over parameter-based methods that suffer from catastrophic forgetting \citep{mquake, EasyEdit, meng2022locating, meng2022mass}. However, existing RAG-based reasoning frameworks exhibit two critical limitations.

First, they are confined to ``knowledge editing''---simple fact replacements \citep{zhang2023large, zhong2023mquake}---and are ill-equipped for \emph{Knowledge Injection}, where models must accommodate both fact updates and entirely novel information \citep{vu2023freshllms}, as in financial news monitoring or scientific literature tracking.

Second, we observe that under Knowledge Injection, hop-level answers become unreliable because the model does not know \emph{which knowledge source to trust}: it over-relies on outdated memory when injected knowledge conflicts with it (K2), and it cannot answer when the fact is entirely new (K3). Existing edited-scenario benchmarks \citep{zhong2023mquake} contain only counterfactual edits, ignoring both the incremental case and this source-trust problem; existing correction mechanisms \citep{zhou2024trustworthiness, yao2023tree} either require training or apply generic heuristics ill-suited to this confidence-based arbitration.

To systematically address these limitations, we make the following contributions. First, we formalize Knowledge Injection by distinguishing two core data types beyond normal knowledge: \textit{conflict data} (K2), which contradicts the model's parametric knowledge, and \textit{incremental data} (K3), which is entirely novel to the model. Second, we construct NEXUS, the first benchmark designed specifically for multi-hop reasoning under Knowledge Injection, explicitly covering both K2 and K3. Third, we propose MIRAGE, an efficient, training-free framework whose core module, the CDR, detects unreliable hop-level answers and resolves them via P(True)-based confidence routing---trusting parametric memory when the model is confident (conflict) and retrieval when it is not (incremental). MIRAGE is complemented by a lightweight Template-Guided Chain Corrector (TGC) that mitigates residual structural deviations (details in Appendix \ref{app:extra_flaws}). Figure \ref{fig:experiments} illustrates the overall architecture.

\begin{figure*}[!t]
\includegraphics[width=\linewidth]{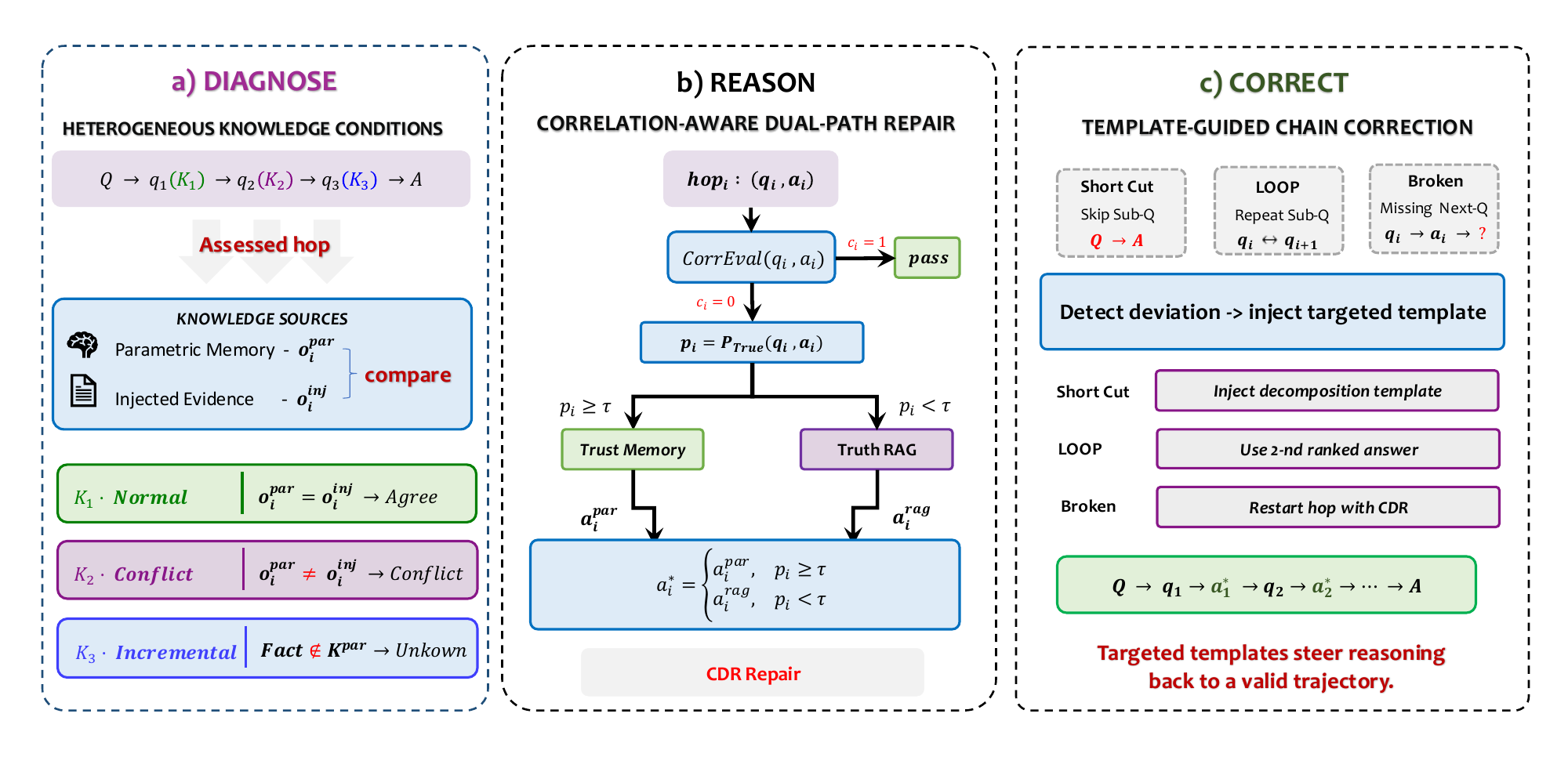}
\caption {Overview of MIRAGE, a training-free framework for reliable multi-hop reasoning under heterogeneous knowledge conditions. (a) Diagnose: MIRAGE characterizes each reasoning hop by comparing parametric memory with injected evidence, distinguishing normal (\(K_1\)), conflict (\(K_2\)), and incremental (\(K_3\)) conditions. (b) Reason: CDR evaluates hop-level answer reliability and routes unreliable answers to parametric-memory or retrieval-based correction according to \(P_{\mathrm{True}}\). (c) Correct: TGC detects shortcut reasoning, reasoning loops, and broken chains, and applies targeted corrections to restore a valid reasoning trajectory.}
\label{fig:experiments}
\end{figure*}

Extensive experiments across four LLM families (LLaMA-2-7B, GPT-3.5-turbo-instruct, Qwen2-7B-Instruct, and GPT-4o) on both MQUAKE-CF-3K and NEXUS demonstrate that MIRAGE consistently outperforms strong baselines including PokeMQA, improving both final answer accuracy and hop-level reasoning accuracy---for instance, on GPT-4o with a single edit, by a 21.5-point gain over PokeMQA on NEXUS---with consistent improvements across varying numbers of edits and model scales.

\section{Related Work}
\subsection{Multi-hop Answering}
MQA requires constructing a reasoning chain over multiple pieces of evidence \citep{yang2018hotpotqa}. Early approaches relied on structured knowledge bases and graph traversal \citep{berant2013semantic,reddy2014large,bordes2014question}; recent LLM-based methods decompose a question into simpler sub-questions and answer them sequentially, e.g., via chain-of-thought or explicit decomposition \citep{dua2022successive, sprague2024cot, chen2022program}. However, most existing frameworks assume a static knowledge environment \citep{meng2022mass, shi2024continual}: they reason over a fixed set of facts and cannot handle dynamically updated knowledge.

\subsection{Knowledge Conflict}
To counter static knowledge, knowledge editing modifies specific facts within a model's parameters, e.g., ROME \citep{meng2022locating} and MEMIT \citep{meng2022mass}, or suppresses undesirable knowledge via unlearning \citep{eldan2310s, yao2024large}. Yet these methods verify single fact updates without checking whether the change propagates through a multi-step chain, and they are not designed for \emph{knowledge injection}---the simultaneous introduction of both updated and entirely new knowledge---which strictly generalizes knowledge editing.

\subsection{Adaptive Retrieval-Augmented Reasoning}

Concurrent work has explored adaptive reasoning for RAG: Self-RAG \citep{asai2024self} uses reflection tokens to trigger retrieval, FLARE \citep{arakelyan2025flare} schedules retrieval by uncertainty, CRAG \citep{yan2024corrective} evaluates retrieved documents and invokes corrective strategies, and ReAct \citep{yao2022react} interleaves reasoning with actions. These methods target open-domain QA over a static knowledge base and either require fine-tuning (Self-RAG) or rely on heuristics (FLARE, CRAG) not tailored to the multi-hop structure.

Most related to our setting, PokeMQA \citep{gu2024pokemqa} interleaves knowledge editing with multi-hop reasoning, but it targets counterfactual edits and does not model incremental knowledge or provide confidence-based source arbitration. In contrast, MIRAGE is \textbf{training-free}, using P(True)-based routing to arbitrate between memory and retrieval, with lightweight template-guided correction for residual structural deviations---to our knowledge, the first systematic study of multi-hop reasoning under knowledge injection with an inference-only correction framework.

\section{The NEXUS Benchmark}
\label{sec:dataset}
\subsection{Knowledge Types}
\begin{figure}[t]
\centering 
\includegraphics[width=0.7\linewidth,clip=true,trim=0 15 0 0]{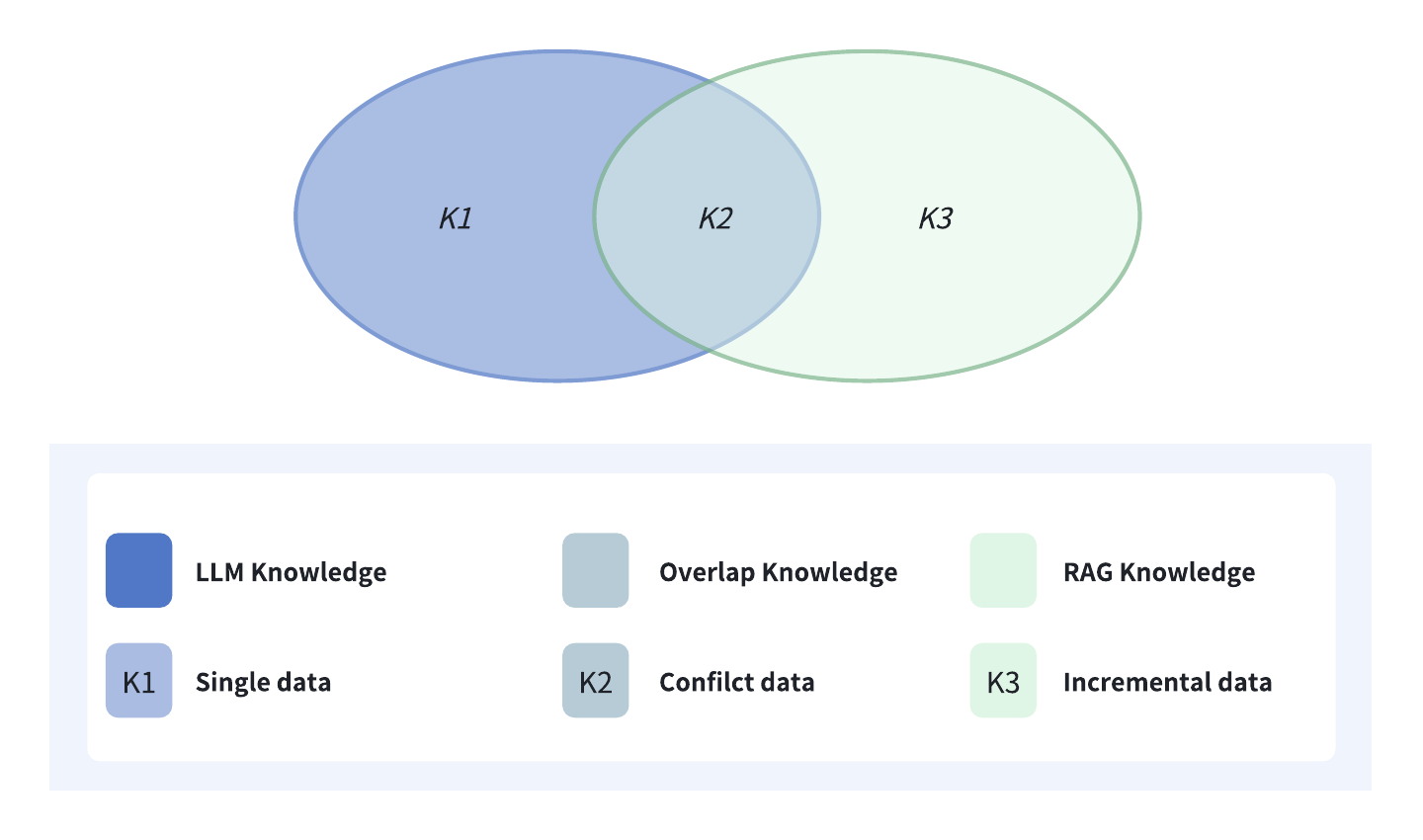}
\caption {A conceptual Venn diagram of the knowledge landscape formed by the LLM's static knowledge and RAG's dynamic knowledge. This landscape is divided into three regions: LLM knowledge (K1), conflict due to outdated LLM knowledge (K2), and new incremental data from RAG (K3). We identify the presence of conflict (K2) and incremental (K3) data as the root cause of reasoning failures. Therefore, our framework is engineered to systematically distinguish these cases, thereby enabling robust and reliable reasoning.}
\label{fig:data}
\end{figure}

To systematically evaluate reasoning under diverse knowledge scenarios, we constructed a dataset categorized by the relationship between the LLM's internal, parametric knowledge and the external knowledge from RAG. We first clarify the core distinction: knowledge editing revises existing facts, while knowledge injection handles both conflicting updates (K2) and novel incremental knowledge (K3). As illustrated in Figure \ref{fig:data}, we distinguish three atomic single-hop data types: \emph{normal data} (K1), \emph{conflict data} (K2), and \emph{incremental data} (K3). Formally, let a multi-hop question $Q$ be decomposed into single-hop sub-questions $\{q_1, \dots, q_n\}$, where $q_i$ targets the fact $(s_i, r_i, o_i)$; let $\mathcal{K}^{\mathrm{par}}$ denote the model's parametric knowledge, and let $o_i^{\mathrm{inj}}$ denote the object supplied by the injected (external) knowledge. Each hop is assigned a type (Eq.~\ref{eq:hop_type})
\begin{equation}
\footnotesize
t_i = \begin{cases}
K1\ \text{(normal)}, & (s_i, r_i) \in \mathcal{K}^{\mathrm{par}}\ \text{and}\ o_i^{\mathrm{inj}} = o_i^{\mathrm{par}},\\
K2\ \text{(conflict)}, & (s_i, r_i) \in \mathcal{K}^{\mathrm{par}}\ \text{and}\ o_i^{\mathrm{inj}} \neq o_i^{\mathrm{par}},\\
K3\ \text{(incremental)}, & (s_i, r_i) \notin \mathcal{K}^{\mathrm{par}},
\end{cases}
\label{eq:hop_type}
\end{equation}
where $o_i^{\mathrm{par}}$ is the object the model recalls for $(s_i, r_i)$.

\begin{figure}[!t]
  \centering
  \includegraphics[width=\linewidth,clip=true,trim=0 0 0 0]{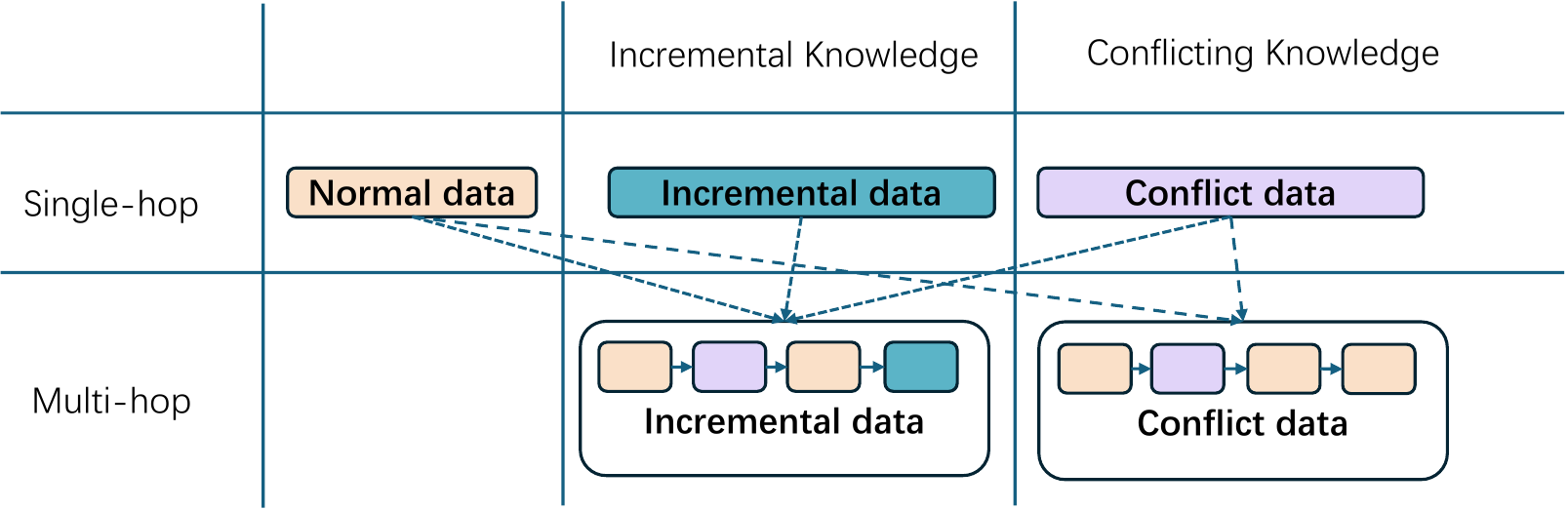}
  \caption{Different single-hop data types combine together to form multi-hop data.}
  \label{app:datatype_fig}
\end{figure}

We extend this classification to the multi-hop level, where a question's category is determined by its most challenging hop (Eq.~\ref{eq:q_type}): a question is \emph{Multi-hop Incremental} if it contains any K3 hop, and otherwise \emph{Multi-hop Conflict} if it includes at least one K2 hop,
\begin{equation}
\footnotesize
T(Q) = \begin{cases}
\text{Incremental}, & \exists\, i : t_i = K3,\\
\text{Conflict}, & \nexists\, i : t_i = K3 \wedge \exists\, i : t_i = K2,\\
\text{Normal}, & \text{otherwise}.
\end{cases}
\label{eq:q_type}
\end{equation}
This reflects the premise that a single instance of conflicting or novel information dictates the entire task's difficulty. The relationship among data types is illustrated in Figure \ref{app:datatype_fig}.

In contrast to existing datasets, which mainly contain trivial factual replacements, NEXUS explicitly models incremental and conflicting knowledge at the atomic hop level and builds multi-hop data from these primitives. 

\subsection{Baseline Geographic Subset}
\label{sec:baseline_geo}
To provide a lightweight and controlled test bed, NEXUS also contains a baseline geographic subset, which serves as the incremental (K3) component of the dataset. Following prior practice, we extracted 461 single-hop country--capital instances from MQUAKE and transformed them into multi-hop incremental instances. For example, the single-hop fact \((s,r,o)=\) (United States, capital, Washington) is converted into a fictional fact \((s',r,o')=\) (Fyldros, capital, Nolvar), yielding the entirely novel question ``What is the capital of Fyldros?''. The full construction procedure, the generation prompt, and representative samples are provided in Appendix \ref{app:fictional_country}.

\subsection{Constructing NEXUS}
\label{sec:dataset_construction}

To scale beyond the baseline subset, we propose an advanced construction pipeline. We collected 4,000 factual items from Wikipedia across six domains (finance, technology, healthcare, sports, economics, education), formulated them into multi-hop questions and knowledge points with GPT-3.5, and obtained an initial set of 3,227 candidate instances.

To cover different knowledge states, we categorize the candidates into \textbf{Incremental Data} and \textbf{Conflict Data}.

\textbf{Incremental Data Construction:} Following the knowledge boundary probing of Ren et al. \citep{ren2025investigating}, we query the target model zero-shot with the generated questions and retain instances the model fails to answer or hallucinates on---the ``unknowns'' missing from its parametric memory.

\textbf{Conflict Data Construction:} For instances the model can answer accurately, we follow the counterfactual paradigm of Longpre et al. \citep{longpre2021entity} and apply an \textbf{Edit-hop} mechanism: using Gemini 2.5 \citep{gemini_team_2023}, we perform counterfactual modifications on intermediate entities within each multi-hop chain, so the model must override its parametric memory.

\textbf{Case Study of the Edit-hop Mechanism:} Consider a two-hop question whose standard reasoning first identifies an intermediate entity and then derives the final answer from it. In the Edit-hop phase, we inject a counterfactual premise into the intermediate hop (e.g., relocating the intermediate entity), forcing the model to override its parametric memory and reason strictly from the injected knowledge. A concrete worked example is provided in Appendix \ref{app:case_study}.

\textbf{Evaluation of the Source-Trust Failure:} This design directly probes the model's ability to arbitrate between parametric memory and injected knowledge at each hop. Empirically, the dominant failure we observe is a \emph{source-trust} failure: when injected knowledge conflicts with parametric memory (K2), the model answers from outdated memory; when the fact is entirely new (K3), it has no memory and hallucinates. Several structural deviations (shortcut reasoning, reasoning loops, broken chains) are secondary; we describe them with simple fix attempts in Appendix \ref{app:extra_flaws}.

After quality filtering of the candidate instances, the final NEXUS dataset comprises 3,100 multi-hop instances: 461 incremental instances, corresponding to the baseline geographic subset of Section \ref{sec:baseline_geo}, and 2,639 conflict instances constructed by the advanced pipeline.

\subsection{Dataset Statistics}
\label{sec:dataset_stats}
The final NEXUS dataset integrates the baseline geographic subset and the cross-domain advanced dataset. At the question level, it contains only instances that involve knowledge injection: 461 multi-hop incremental (K3) instances and 2,639 multi-hop conflict (K2) instances, for 3,100 in total. At the single-hop level, the three knowledge types follow a ratio of approximately 5:3:2, with normal (K1) hops as the main body, followed by conflict (K2) and incremental (K3) hops. In terms of hop distribution, the dataset is dominated by 2-hop instances (about 60\%), supplemented by 3-hop (about 30\%) and 4-hop (about 10\%) instances, balancing task rationality and challenge. The fused dataset covers seven domains in total---the six cross-domain Wikipedia categories (finance, technology, healthcare, sports, economics, and education) plus the geographic subset---with each domain's proportion controlled between 10\% and 20\%.

\section{The MIRAGE Framework}
\label{sec:mirage_framework}
\begin{figure*}[t]
\includegraphics[width=\linewidth,clip=true,trim=0 0 0 0]{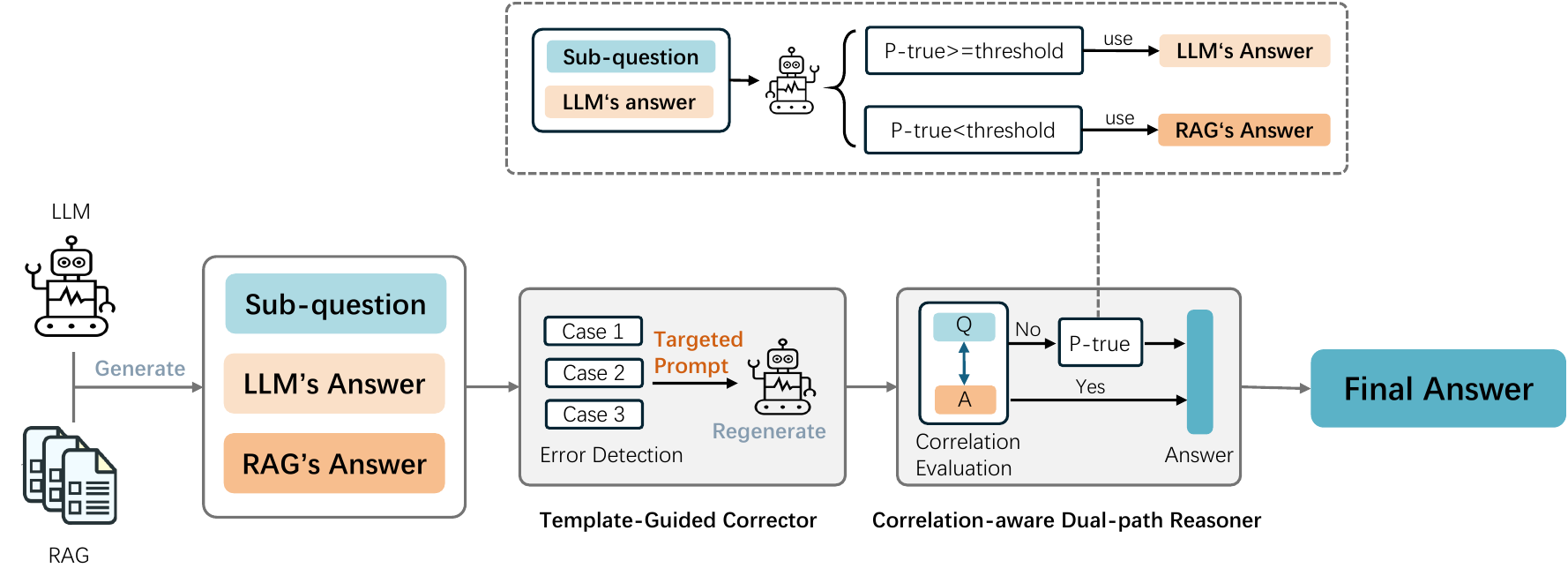}
\vspace{1ex}
\caption {Step-by-step execution workflow of the MIRAGE framework (one hop of Algorithm \ref{alg:mirage}). The CDR (core) verifies each hop's answer against its sub-question via the correlation evaluator; on inconsistency ($c_i=0$), the P(True) routing decides whether to trust parametric memory (conflict, K2) or retrieval (incremental, K3). The TGC (supporting) detects and mitigates structural deviations in the reasoning path.}
\label{fig:main_2}
\end{figure*}

Figure \ref{fig:main_2} illustrates the structure of MIRAGE. The framework is organized around a core module---the CDR---which corrects the dominant failure under Knowledge Injection: a hop-level answer that trusts the wrong knowledge source. At each hop, MIRAGE retrieves evidence, generates a candidate answer, and uses the CDR to check whether the candidate is consistent with its sub-question; if not, a P(True)-based routing rule (Section \ref{sec:cdr}) decides whether to trust parametric memory or retrieval. A lightweight TGC supports the process by detecting residual structural deviations in the reasoning path; its details and simple fixes are given in Appendix \ref{app:extra_flaws}. Algorithm \ref{alg:mirage} summarizes the procedure.

\begin{algorithm}[t]
\caption{MIRAGE: multi-hop reasoning under Knowledge Injection.}
\label{alg:mirage}
\begin{algorithmic}[1]
\STATE \textbf{Input:} question $Q$, retriever $\mathcal{R}$, base LLM $\mathcal{M}$, threshold $\tau=0.5$.
\STATE \textbf{Output:} final answer $A$.
\STATE Decompose $Q$ into sub-questions $\{q_1,\dots,q_n\}$ using $\mathcal{M}$.
\FOR{$i=1$ \textbf{to} $n$}
    \STATE Retrieve $e_i\leftarrow\mathcal{R}(q_i)$; generate $a_i\leftarrow\mathcal{M}(q_i,e_i)$.
    \STATE If a structural deviation is detected, regenerate $q_i$ or $a_i$. \COMMENT{TGC}
    \STATE Evaluate $c_i\leftarrow\mathrm{CorrEval}(q_i,a_i)$. \COMMENT{CDR}
    \IF{$c_i=0$}
        \STATE Estimate $p_i\leftarrow P_{\mathrm{True}}(q_i,a_i)$.
        \IF{$p_i\ge\tau$}
            \STATE $a_i\leftarrow a_i^{\mathrm{par}}$. \COMMENT{conflict: trust memory}
        \ELSE
            \STATE $a_i\leftarrow a_i^{\mathrm{rag}}$. \COMMENT{incremental: trust retrieval}
        \ENDIF
    \ENDIF
\ENDFOR
\STATE Compose final answer $A$ from $\{a_1,\dots,a_n\}$.
\STATE \textbf{return} $A$.
\end{algorithmic}
\end{algorithm}

\subsection{Correlation-aware Dual-path Reasoner (CDR)}
\label{sec:cdr}
The CDR is the core module of MIRAGE. It comprises a correlation evaluator, which detects unreliable hop-level answers, and the CDR core, which repairs them by choosing the right knowledge source. 
\subsubsection{Correlation Evaluator}
The correlation evaluator serves as the gatekeeper of the CDR. Its role is to assess whether the retrieved answer $a_i$ is consistent with its sub-question $q_i$. To avoid introducing external biases, we use the same base LLM to judge the semantic alignment; the full prompt template is provided in Appendix \ref{app:correlation_prompt}. The evaluation yields a binary signal (Eq.~\ref{eq:corr})
\begin{equation}
c_i = \mathrm{CorrEval}(q_i, a_i) \in \{0, 1\},
\label{eq:corr}
\end{equation}
where $c_i = 1$ indicates consistency, in which case the hop is accepted without correction, and $c_i = 0$ triggers the CDR core.

\subsubsection{CDR Core}
The primary role of the CDR core is to diagnose why a hop-level answer is unreliable and to select the appropriate source to trust. The key distinction is whether the model retains prior parametric knowledge of the queried fact: for conflict data (K2) the model is typically confident about its (outdated) memory, whereas for incremental data (K3) it has no memory and is uncertain. Among uncertainty estimators such as Monte-Carlo Dropout \citep{MC16} and Deep Ensembles \citep{DE17}, we adopt the P(True) self-reflection estimator \citep{ptrue22Ka, ptrue24nature}, which reformulates the decision into a single binary-answer query and is therefore training-free and cheaper than token-level logit-based evaluations \citep{uncer20Malinin}. Given a sub-question and its candidate answer, the CDR core estimates (Eq.~\ref{eq:ptrue})
\begin{equation}
p_i = P_{\mathrm{True}}(q_i, a_i) \in [0, 1],
\label{eq:ptrue}
\end{equation}
using the prompt in Appendix \ref{app:ptrue_prompt}, and applies the routing rule (Eq.~\ref{eq:routing})
\begin{equation}
a_i^* = \begin{cases}
a_i^{\mathrm{par}}, & p_i \ge \tau \quad \text{(conflict)},\\
a_i^{\mathrm{rag}}, & p_i < \tau \quad \text{(incremental)},
\end{cases}
\label{eq:routing}
\end{equation}
where $\tau = 0.5$ by default, $a_i^{\mathrm{par}}$ is the model's own answer, and $a_i^{\mathrm{rag}}$ is obtained by re-answering from the RAG evidence. Intuitively, high confidence indicates that the error stems from outdated memory being favored over injected knowledge (K2), so we retain the model's answer as the single-hop resolution; low confidence indicates that the fact lies beyond the model's knowledge (K3), so we trust retrieval instead.

\subsection{Template-Guided Chain Corrector (TGC)}
Beyond the per-hop source-trust errors addressed by the CDR, we additionally observe a small set of structural deviations in the reasoning process---shortcut reasoning (answering without decomposing), reasoning loops (repeatedly producing the same sub-question), and broken chains (failing to generate the next sub-question). Because these deviations are lightweight and recur across a few recognizable patterns, we handle them with the TGC, a zero-training module that recognizes the deviation and injects a targeted template to steer the model back on track (e.g., enforcing step-by-step decomposition). We give the full description of these deviations and our simple fix attempts in Appendix \ref{app:extra_flaws}; the decomposition template is provided in Appendix \ref{app:tgc_prompt}.

\section{Experiments}
\label{sec:experiments}
\subsection{Experimental Setup}
\label{sec:setup}
We evaluate MIRAGE on four base LLMs spanning different families and scales: LLaMA-2-7B, GPT-3.5-turbo-instruct, Qwen2-7B-Instruct, and GPT-4o. We compare against parameter-based knowledge-editing baselines (MEMIT, MELLO) and the state-of-the-art multi-hop reasoning-under-edits method PokeMQA. Because MEMIT and MELLO require direct access to model parameters, they are reported on the open-weight LLaMA-2-7B backbone; PokeMQA serves as the primary comparison baseline on all backbones. Following prior work, we consider three edit settings---\textit{1 edited}, \textit{100 edited}, and \textit{All edited}---and retrieve $k$ documents per reasoning step. We report two metrics: final answer accuracy (Acc) and hop-level reasoning accuracy (Hop-Acc). Full implementation and evaluation details, including hyperparameters and compute resources, are provided in Appendix \ref{app:impl}.

\begin{table*}[!t]
\centering
\caption{Evaluation results on MQUAKE-CF-3K and NEXUS. All metrics are reported in percentage (\%).}
\label{tab:maintable}
\resizebox{\textwidth}{!}{
\begin{tabular}{@{} l cccccc cccccc @{}}
\toprule
\multirow{3}{*}{\textbf{Method}} & \multicolumn{6}{c}{\textbf{MQUAKE-CF-3K}} & \multicolumn{6}{c}{\textbf{NEXUS}} \\
\cmidrule(lr){2-7} \cmidrule(l){8-13}
& \multicolumn{2}{c}{\textit{1 edited}} & \multicolumn{2}{c}{\textit{100 edited}} & \multicolumn{2}{c}{\textit{All edited}} 
& \multicolumn{2}{c}{\textit{1 edited}} & \multicolumn{2}{c}{\textit{100 edited}} & \multicolumn{2}{c}{\textit{All edited}} \\
\cmidrule(lr){2-3} \cmidrule(lr){4-5} \cmidrule(lr){6-7} \cmidrule(lr){8-9} \cmidrule(lr){10-11} \cmidrule(l){12-13}
& Acc & Hop-Acc & Acc & Hop-Acc & Acc & Hop-Acc & Acc & Hop-Acc & Acc & Hop-Acc & Acc & Hop-Acc \\
\midrule

\multicolumn{13}{@{}l}{\textbf{LLaMA-2-7B}} \\
\midrule
MEMIT                    & 14.97 & 6.43  & 9.4   & 2.47  & 2.3   & 0.37  & 13.82 & 4.92  & 8.26  & 3.04  & 1.8   & 1.2   \\
MELLO                    & 33.57 & 9.9   & 20.0  & 10.07 & 17.33 & 9.91  & 28.29 & 9.87  & 18.42 & 8.7   & 14.92 & 7.82  \\
PokeMQA                  & 44.13 & 30.6  & 37.33 & 27.83 & 32.83 & 23.87 & 42.27 & 28.86 & 34.26 & 27.12 & 32.12 & 21.12 \\
MIRAGE (Ours)            & \textbf{49.72} & 31.8  & \textbf{41.12} & 30.36 & \textbf{38.1}  & 26.44 & \textbf{42.67} & 31.21 & \textbf{36.52} & 28.51 & \textbf{33.15} & 22.41 \\
\midrule

\multicolumn{13}{@{}l}{\textbf{GPT-3.5-turbo-instruct}} \\
\midrule
PokeMQA                  & 61.28 & 56.12 & 55.78  & 47.32 & 43.19 & 36.51 & 59.12 & 51.23 & 54.2  & 45.77& 47.69 & 40.22 \\
MIRAGE (Ours)            & \textbf{68.41} & 57.21 & \textbf{57.83} & 50.15 & \textbf{45.22} & 38.20 & \textbf{64.22} & 52.20 & \textbf{54.67} & 47.31 & \textbf{49.21} & 42.20 \\
\midrule

\multicolumn{13}{@{}l}{\textbf{Qwen2-7B-Instruct}} \\
\midrule
PokeMQA                  & 66.39 & 30.23 & 42.19 & 24.48 & 40.01 & 22.55 & 66.13 & 31.94 & 40.55 & 23.18 & 37.62 & 21.16 \\
MIRAGE (Ours)            & \textbf{73.28} & 35.55 & \textbf{47.83} & 26.02 & \textbf{45.72} & 23.22 & \textbf{70.42} & 32.36 & \textbf{43.15} & 24.64 & \textbf{38.15} & 22.45 \\
\midrule

\multicolumn{13}{@{}l}{\textbf{GPT-4o}} \\
\midrule
PokeMQA                  & 63.65 & 52.50 & 41.44 & 39.44 & 39.75 & 37.98 & 52.15 & 51.80 & 44.23 & 43.87 & 38.70 & 37.11 \\
MIRAGE (Ours)            & \textbf{78.29} & 59.21 & \textbf{47.98} & 45.54 & \textbf{42.25} & 40.28 & \textbf{73.65} & 63.43 & \textbf{46.98} & 45.37 & \textbf{40.71} & 39.02 \\
\bottomrule
\end{tabular}
}
\vspace{0.5ex}

\begin{minipage}{\textwidth}
\footnotesize
\raggedright
\textit{Note:} MEMIT and MELLO require direct access to model parameters and are reported only on the open-weight LLaMA-2-7B backbone.
\end{minipage}
\end{table*}

Table \ref{tab:maintable} reports results across four base LLMs and two datasets. MIRAGE consistently outperforms all baselines, including PokeMQA, in both final-answer accuracy (Acc) and hop-level reasoning accuracy (Hop-Acc). The gains are consistent across all four backbones: on MQUAKE-CF-3K (1 edit), for instance, Acc rises from 61.28 to 68.41 (+7.13) on GPT-3.5-turbo-instruct and reaches 73.28 on Qwen2-7B-Instruct, while on NEXUS MIRAGE gains a notable 21.5 points on GPT-4o (52.15 to 73.65). These results confirm that the CDR's confidence-based source-trust correction is effective across edit settings, and that MIRAGE also strengthens the reasoning chain itself, improving Hop-Acc in most configurations, e.g., GPT-3.5-turbo-instruct (56.12 to 57.21, 1 edit) and on NEXUS for LLaMA-2-7B, Qwen2-7B-Instruct, and GPT-4o.

\subsection{Analysis of P(True)}

We validate the P(True) metric's ability to separate the two data types on NEXUS: Table \ref{tab:ptrue_stats} reports the mean and variance per category, and the full distributions are plotted in Appendix \ref{app:ptrue_fig}. The results reveal a clear dichotomy---incremental data concentrates near 0.0 while conflict data clusters near 1.0---confirming that P(True) is an effective and reliable signal for our dual-path routing.

\begin{table}
\centering
\caption{Comparison of P(True) statistics for Incremental and Conflict data.}
\vspace{1ex}
\setlength{\tabcolsep}{0.pt}
\label{tab:ptrue_stats}
\begin{tabular*}{\columnwidth}{@{\extracolsep{\fill}} l c c @{}}
\toprule
\textbf{Metric} & \textbf{Incremental data} & \textbf{Conflict data} \\
\midrule
Mean            & 0.24                      & 0.77                   \\
Variance        & 0.0266                     & 0.0976                  \\
\bottomrule
\end{tabular*}
\end{table}

We set 0.5 as the default threshold, though it can be adjusted based on specific tasks. A higher threshold risks misclassifying conflict data as incremental, while a lower threshold may misidentify incremental data as conflict. We additionally validate the correlation evaluator component on the HotpotQA development set and MQUAKE-CF-3K in Appendix \ref{app:corr_eval}.

\subsection{Ablation Study}

Table \ref{tab:gpt35} isolates the contribution of each module on GPT-3.5-turbo-instruct. Used alone, the CDR improves Acc on MQUAKE-CF-3K to 64.01 (1 edit) and 44.08 (All edits), and on NEXUS to 63.23 and 48.46, confirming that source-trust routing enhances accuracy. Used alone, the TGC boosts Acc from 61.28 to 67.94 on MQUAKE-CF-3K (1 edit) and to 63.57 and 49.01 on NEXUS, confirming that addressing structural deviations effectively complements the CDR. Combining both modules yields the best results on every setting (68.41, 45.22, 64.22, and 49.21).

\begin{table}
\centering
\footnotesize
\captionsetup{width=\columnwidth}
\setlength{\tabcolsep}{2pt}
\caption{Ablation Study on GPT-3.5-turbo-instruct.}
\label{tab:gpt35}
\begin{tabular*}{\columnwidth}{@{\extracolsep{\fill}} l cc cc cc @{}}
\toprule
\multicolumn{3}{@{}l}{\textbf{Modules}}
    & \multicolumn{2}{c}{\textbf{MQUAKE-CF-3K}}
    & \multicolumn{2}{c}{\textbf{NEXUS}} \\
\cmidrule(lr){4-5} \cmidrule(l){6-7}
\textbf{Method} & \textbf{CDR} & \textbf{TGC}
    & \textit{1 ed.} & \textit{All ed.}
    & \textit{1 ed.} & \textit{All ed.} \\
\midrule
PokeMQA   & --         & --         & 61.28          & 43.19          & 59.12          & 47.69          \\
\quad+TGC & --         & \checkmark & 67.94          & 45.21          & 63.57          & 49.01          \\
\quad+CDR & \checkmark & --         & 64.01          & 44.08          & 63.23          & 48.46          \\
Full      & \checkmark & \checkmark & \textbf{68.41} & \textbf{45.22} & \textbf{64.22} & \textbf{49.21} \\
\bottomrule
\end{tabular*}
\end{table}

\subsection{MQuAKE-Remastered}
To verify MIRAGE's generalization beyond the primary benchmarks, we further evaluate it on the more recent MQuAKE-Remastered benchmark \citep{zhong2024mquake} using GPT-4o. The full results and analysis are reported in Appendix \ref{app:mquake}.

\subsection{Approaching the Theoretical Upper Bound}
\label{sec:upper_bound}
To assess how close MIRAGE comes to an ideal setting, we evaluate a pure large language model (Qwen2-7B-Instruct) on the unedited MQUAKE dataset as a reference upper bound; here, $k$ denotes the number of documents retrieved per reasoning step. As shown in Table \ref{tab:accents_formatted}, this ``oracle'' ceiling is 76.10 Acc (at $k=1$).

On the edited MQUAKE-CF-3K, our framework achieves 73.28 Acc (at $k=1$), recovering up to 96\% of the unedited-data reference performance (73.28 / 76.10) despite operating under knowledge injection.

The accuracy drops at $k=100$ (47.83) because more retrieved documents raise the probability of RAG matching errors; in multi-hop reasoning, a single incorrect hop can derail the whole chain. MIRAGE's robustness in this high-noise setting underscores its resilience.

\begin{table}[!t]
\centering
\small
\captionsetup{width=\columnwidth}
\setlength{\tabcolsep}{4pt}
\caption{Comparison with Unedited-Data Reference Performance.}
\label{tab:accents_formatted}
\begin{tabular*}{\columnwidth}{@{\extracolsep{\fill}} ll cc @{}}
\toprule
\textbf{Dataset} & \textbf{Method} & \textbf{$k{=}1$} & \textbf{$k{=}100$} \\
\midrule
MQUAKE (no edits) & pure LLM & 76.10 & 72.42 \\
MQUAKE-CF-3K      & Ours     & 73.28 & 47.83 \\
\bottomrule
\end{tabular*}

\begin{minipage}{\columnwidth}
\footnotesize
\raggedright
\textit{Note:} For the edited dataset, the number of edited facts equals $k$ per reasoning step (1 edited $\leftrightarrow k{=}1$,100 edited $\leftrightarrow$ $ k{=}100$), aligning with Table~\ref{tab:maintable}.
\end{minipage}
\end{table}

Qualitative case studies of reasoning failures and their correction under Knowledge Injection are provided in Appendix \ref{app:case_study}; Appendix \ref{app:financial_case} additionally shows a real-world financial instance in which compounding counterfactual premises trigger a complete ``Reasoning Break''.


\section{Conclusion and Future Work}
We identify the knowledge injection problem in multi-hop reasoning and introduce NEXUS, the first benchmark tailored for this setting, which systematically distinguishes conflict data (K2) from incremental data (K3). Under injection, multi-hop reasoning fails at the level of individual hops because the model trusts the wrong knowledge source. To address this, we propose MIRAGE, a training-free framework whose core, the CDR, resolves such source-trust errors via P(True)-based confidence routing, complemented by a lightweight TGC for residual structural deviations. Experiments across four LLM families and two benchmarks show that MIRAGE consistently improves both final-answer and hop-level reasoning accuracy, and reaches approximately 96\% of the unedited-data reference accuracy.

Our analysis also suggests several promising directions for future work. First, the P(True) routing currently relies on a fixed threshold; automatic, task-adaptive threshold selection could further improve robustness. Second, extending NEXUS to non-English and specialized domains (e.g., medical and legal reasoning) would test the generality of our findings. Third, coupling MIRAGE with retrieval-quality estimation could mitigate the risk of propagating erroneous retrieved content. We hope this work encourages further exploration of robust multi-hop reasoning in dynamically evolving knowledge environments.

\section*{Limitations}
MIRAGE relies on existing RAG systems, so retrieval quality remains a key factor in overall performance. Our relevance matching and synonym substitution strategies improve matching without changing the retriever, but stronger retrieval could further improve the framework.
The incremental data are still generated with self-supervised LLMs, and future work could improve quality with human annotation or stronger knowledge-injection pipelines. In addition, the current template-guided corrector only addresses single-hop corrections and could be extended to more complex reasoning chains.




\bibliography{custom}

\appendix
%

\section{Dataset Construction Details}
\label{app:dataset}

\subsection{Fictional Country Generation for the Baseline Geographic Subset}
\label{app:fictional_country}

The baseline geographic subset is built from 461 single-hop country--capital
instances extracted from MQUAKE \citep{mquake}. To convert them into
multi-hop \emph{incremental} instances, we first rewrite single-hop facts into
fictional facts. For example, the single-hop fact
\(e=(s_i,r_i,o_i)=\text{(United States, capital, Washington)}\) is transformed
into \(e'=(s'_i,r_i,o'_i)=\text{(Fyldros, capital, Nolvar)}\), which gives rise
to the entirely novel question ``What is the capital of Fyldros?'' because the
LLM has no prior knowledge of the fictional entities.

We used GPT-3.5-turbo to generate fictional countries together with their
corresponding capitals, and verified the generated data to remove duplicates.
The prompt template used for generation is shown below.

\begin{tcolorbox}[
    colback=yellow!20,
    colframe=yellow!50!black,
    coltitle=black,
    rounded corners,
    boxrule=0.5mm,
    width=\linewidth,
    arc=4mm,
    auto outer arc,
    sharp corners=south,
    before skip=5pt,
    after skip=5pt,
    enhanced
]
Please generate a fictional country and its corresponding fictional capital. The template should be: country: Faldros, capital: Thalnor. Generate 500 data. Please pay attention to capitalization. There is no need to generate sequence numbers.
\end{tcolorbox}

A selection of representative examples from the generated data is presented in
Table \ref{tab:fictional_data_samples}. To integrate such incremental data
into a multi-hop reasoning chain, for the hop that precedes the incremental
hop, \(e=(s_{i-1},r_{i-1},o_{i-1})\), we edit its object to
\(o_{i-1}^*=s_i'\). Since incremental data typically occur at the final hop,
no further integration is required.

\begin{table}[t]
\centering
\caption{Samples of the generated fictional countries and their corresponding capitals.}
\label{tab:fictional_data_samples}
\renewcommand{\arraystretch}{1.1}
\begin{tabular}{ll @{\hspace{4em}} ll}
\toprule
\textbf{Country} & \textbf{Capital} & \textbf{Country} & \textbf{Capital} \\
\midrule
Faldros  & Thalnor  & Arthos   & Olmara  \\
Aleria   & Vendrell & Sylvoris & Trenna  \\
Drathos  & Harnath  & Tarlen   & Maznor  \\
Siltara  & Zorvona  & Hytrion  & Vortum  \\
Malira   & Bryntor  & Belithar & Lothar  \\
Velora   & Theralis & Elarion  & Firos   \\
\bottomrule
\end{tabular}
\end{table}


\section{Prompt Templates}
\label{app:prompts}

This section lists the complete prompt templates used by the components of the
MIRAGE framework. All prompts are applied in a zero-shot, training-free manner
with the same base LLM.

\subsection{Correlation Evaluator}
\label{app:correlation_prompt}

The correlation evaluator judges whether a retrieved answer is semantically
aligned with its sub-question. The prompt is:

\begin{tcolorbox}[
    colback=yellow!20,
    colframe=yellow!50!black,
    coltitle=black,
    rounded corners,
    boxrule=0.5mm,
    width=\linewidth,
    arc=4mm,
    auto outer arc,
    sharp corners=south,
    enhanced
]
Task: Determine if the Answer is relevant to the Original Question.
An answer is relevant if it directly addresses the question.

Respond with only ``Related'' or ``Unrelated''.

---

\textbf{Original Question:} \{original\_question\} \par
\textbf{Answer:} \{sub\_question\} \par
\vspace{1em}
\textbf{Judgment:}
\end{tcolorbox}

\subsection{P(True) Confidence Estimation}
\label{app:ptrue_prompt}

P(True) \citep{ptrue22Ka, ptrue24nature} reformulates the model's response into
a binary classification task to estimate the probability that a proposed answer
is correct. The prompt is:

\begin{tcolorbox}[
    colback=yellow!20,
    colframe=yellow!50!black,
    coltitle=black,
    rounded corners,
    boxrule=0.5mm,
    width=\linewidth,
    arc=4mm,
    auto outer arc,
    sharp corners=south,
    enhanced
]
Question: ``Where is the hometown of Ne Zha?''

LLM: ``China.''

Is the proposed answer correct? Please choose from the following options:

A. True

B. False
\end{tcolorbox}

\subsection{Template-Guided Chain Corrector}
\label{app:tgc_prompt}

When the TGC detects the \emph{Overconfident Reasoning} flaw (i.e., the model
skips sub-question decomposition and directly emits a final answer), it injects
the following template to enforce step-by-step decomposition:

\begin{tcolorbox}[
    colback=yellow!20,
    colframe=yellow!50!black,
    coltitle=black,
    rounded corners,
    boxrule=0.5mm,
    width=\linewidth,
    arc=4mm,
    auto outer arc,
    sharp corners=south,
    enhanced
]
Please revise your above answer according to the template:

You need to answer a multi-hop question.

You should follow the rules below step by step:

1. First, decompose the question into simple subquestions.

2. Then, answer each subquestion.

3. Finally, give the final answer.

You must use the following template:

Subquestion:

Generated answer:

Final answer:
\end{tcolorbox}

\section{Additional Structural Failure Modes and Simple Fixes}
\label{app:extra_flaws}

Beyond the core source-trust failure addressed by the CDR
(Section \ref{sec:cdr}), we additionally observe three recurring structural
deviations in the multi-hop reasoning process under Knowledge Injection. These
are secondary to the source-trust problem, and we make lightweight fix attempts
for each of them.

\begin{itemize}
    \item \textbf{Shortcut reasoning (overconfidence).} The model skips
    sub-question decomposition and directly emits a final answer, ignoring the
    injected knowledge. In our analysis this is the most frequent structural
    deviation (42.3\%). Our fix is to inject a template that enforces
    step-by-step decomposition (Section \ref{app:tgc_prompt}).
    \item \textbf{Reasoning loops (repetitive matching).} Distinct sub-questions
    collapse onto the same retrieved answer, so the chain cannot progress
    (18.8\%). Our fix is to replace the repeated answer with the
    second-highest-scoring retrieval result.
    \item \textbf{Broken chains (missing sub-questions).} The model fails to
    generate the next logically connected sub-question (28.6\%). Our fix is to
    invoke the CDR's relevance detection to obtain a correct retrieval result
    and restart the next sub-question.
\end{itemize}

We emphasize that these fixes are lightweight complements; the main corrective
mechanism of MIRAGE is the CDR's confidence-based source-trust routing.

\section{Implementation and Experimental Details}
\label{app:impl}

\subsection{Models and Baselines}
We evaluate MIRAGE on four base LLMs spanning different families and scales:
LLaMA-2-7B \citep{touvron2023llama}, GPT-3.5-turbo-instruct, Qwen2-7B-Instruct
\citep{team2024qwen2}, and GPT-4o \citep{openai2024gpt4o}. We compare against the parameter-based
knowledge-editing baselines MEMIT \citep{meng2022mass} and MELLO, as well as
the state-of-the-art multi-hop reasoning-under-edits method PokeMQA
\citep{gu2024pokemqa}.

\subsection{Evaluation Protocol}
Following prior work, we consider three edit settings on each benchmark:
\textit{1 edited}, \textit{100 edited}, and \textit{All edited}. For each
reasoning step, $k$ documents are retrieved from the RAG index; we report
results for $k=1$ and $k=100$. Two metrics are reported:
\begin{itemize}
    \item \textbf{Acc}: the accuracy of the final answer.
    \item \textbf{Hop-Acc}: the hop-level reasoning accuracy, i.e., the
    proportion of reasoning hops whose intermediate answers are correct.
\end{itemize}
The correlation evaluator is additionally evaluated on the HotpotQA
development set \citep{yang2018hotpotqa} (7,405 instances) and on
MQUAKE-CF-3K \citep{mquake} (3,000 instances).

\subsection{Hyperparameters}
The P(True) routing threshold is set to 0.5 by default. No fine-tuning is
performed for any model; MIRAGE operates purely at inference time.

\subsection{Compute Resources}
MIRAGE is a training-free framework that requires no additional model training. The experiments were conducted using NVIDIA V100 GPUs, with computational costs primarily arising from LLM inference, retrieval, and hop-level verification.

\section{Additional Experimental Results}
\label{app:additional}

\subsection{P(True) Distribution}
\label{app:ptrue_fig}
\begin{figure}[!t]
\centering
\begin{subfigure}[b]{0.48\columnwidth}
    \centering
    \includegraphics[width=\linewidth]{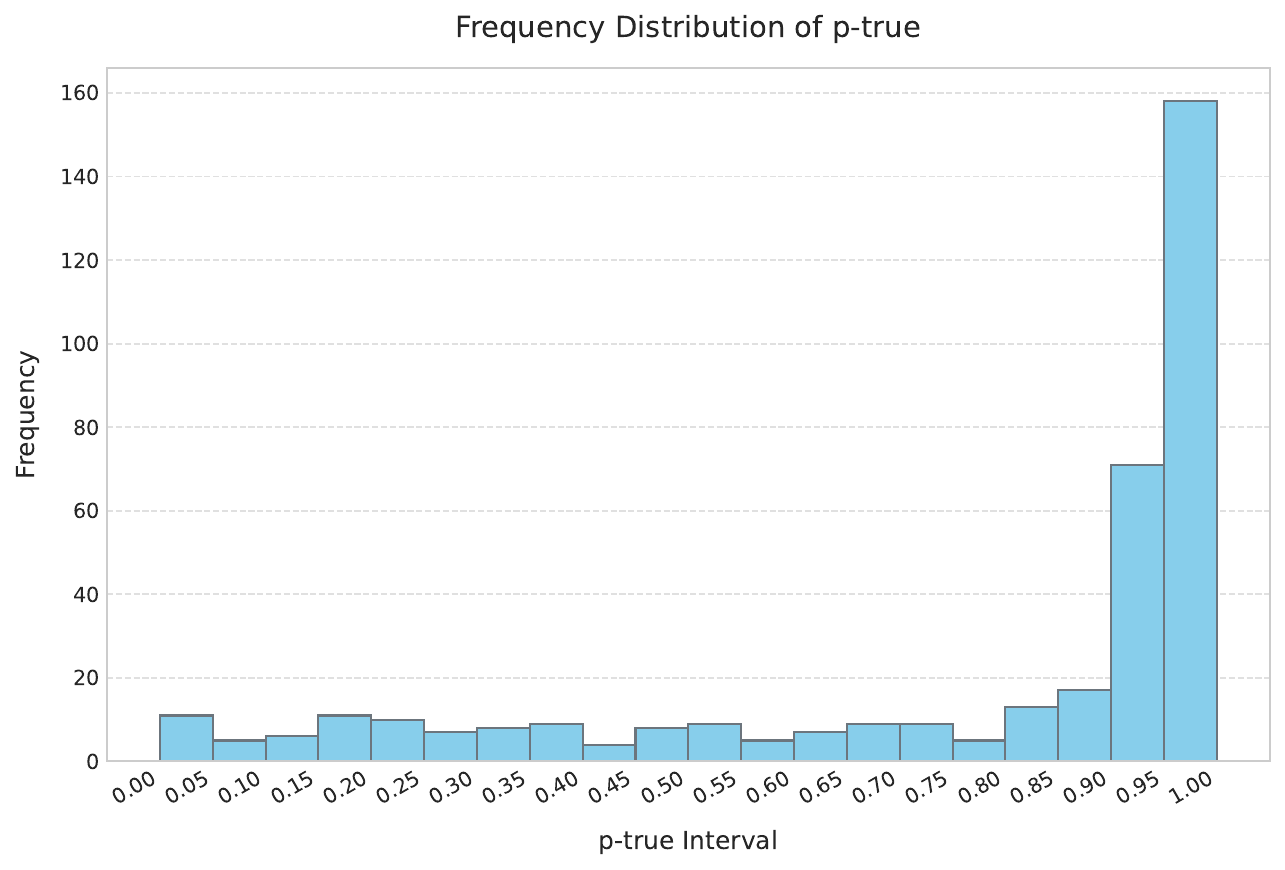}
    \caption{Conflict data}
    \label{fig:dist_conflict}
\end{subfigure}
\hfill
\begin{subfigure}[b]{0.48\columnwidth}
    \centering
    \includegraphics[width=\linewidth]{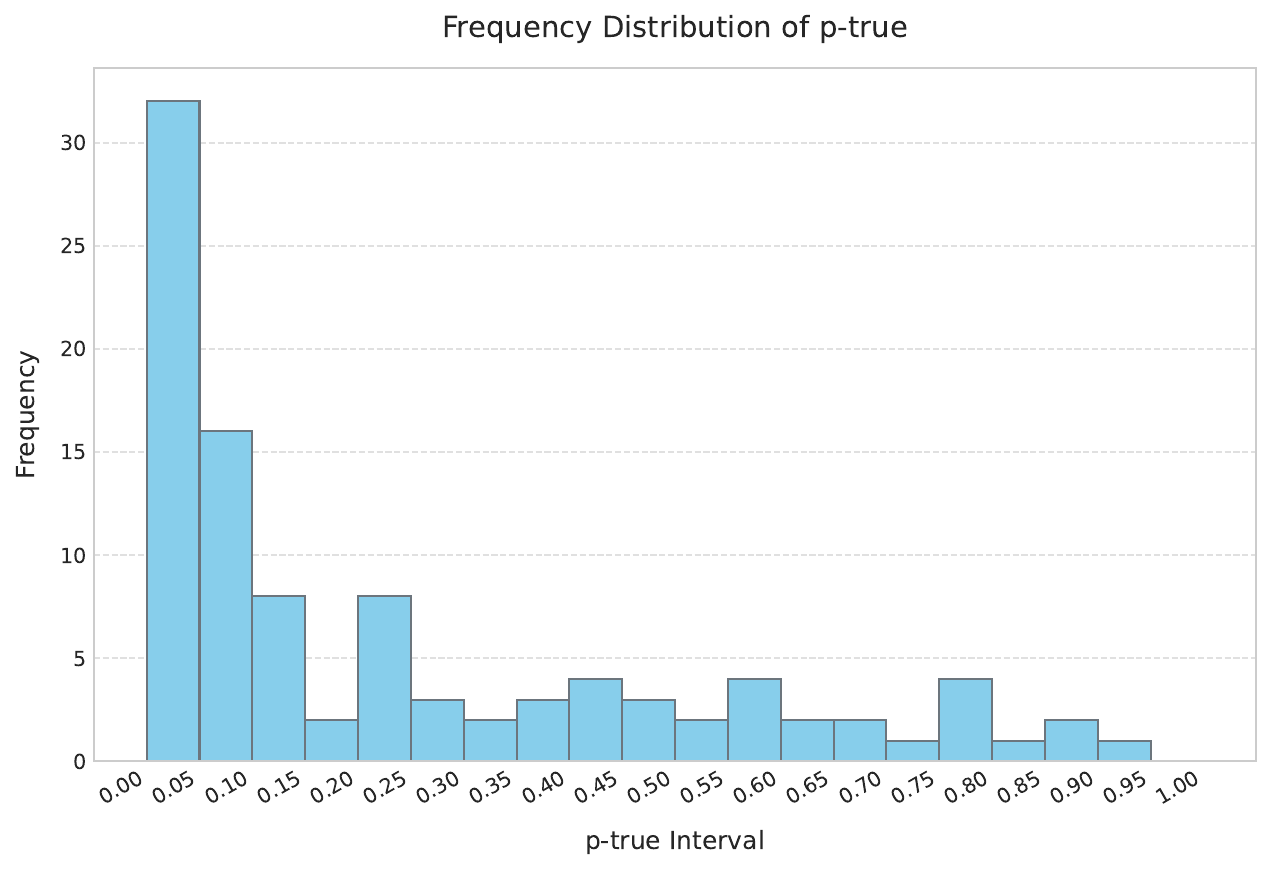}
    \caption{Incremental data}
    \label{fig:dist_incremental}
\end{subfigure}
\caption{Frequency distribution of P(True). (a) Conflict data exhibits high confidence scores, whereas (b) incremental data is concentrated at lower confidence levels. Based on this contrast, 0.5 is recommended as the default threshold.}
\label{fig:distribution}
\end{figure}

Figure \ref{fig:distribution} plots the full P(True) distributions for
conflict and incremental data, confirming the clear separation used by the
CDR's routing rule.

\subsection{Evaluation on MQuAKE-Remastered}
\label{app:mquake}

During manuscript preparation, the optimized MQuAKE-Remastered
\citep{zhong2024mquake} was released as an updated MQuAKE variant. To verify
MIRAGE's generalization, we conducted consistent experiments on this dataset,
comparing our method with the baseline method on GPT-4o. As shown in
Table \ref{tab:gpt4o_supplementary_tblr_small}, MIRAGE demonstrates
significant advantages. In the \textit{1 edited} scenario, it boosts Acc from
36.29 to 54.77 (+18.48) and Hop-Acc from 32.69 to 53.64 (+20.95). Even in the
challenging \textit{100 edited} setting, MIRAGE maintains a consistent lead in
both Acc (23.95 vs.\ 21.60) and Hop-Acc (21.91 vs.\ 20.23), confirming its
robust generalization.

\begin{table}[!t]
\centering
\small
\captionsetup{width=\columnwidth}
\setlength{\tabcolsep}{3pt}
\caption{Evaluation results on the MQuAKE-Remastered dataset (\%).}
\label{tab:gpt4o_supplementary_tblr_small}
\begin{tabular*}{\columnwidth}{@{\extracolsep{\fill}} l cccc @{}}
\toprule
\multirow{2}{*}{\textbf{Method}} & \multicolumn{2}{c}{\textbf{1 edited}} & \multicolumn{2}{c}{\textbf{100 edited}} \\
\cmidrule(lr){2-3} \cmidrule(l){4-5}
& Acc & Hop-Acc & Acc & Hop-Acc \\
\midrule
PokeMQA       & 36.29          & 32.69          & 21.60          & 20.23          \\
MIRAGE (Ours) & \textbf{54.77} & \textbf{53.64} & \textbf{23.95} & \textbf{21.91} \\
\bottomrule
\end{tabular*}
\end{table}

\subsection{Correlation Evaluator Validation}
\label{app:corr_eval}
To assess the correlation evaluator thoroughly, we evaluate it on two distinct datasets, each chosen for a specific purpose. First, we use the HotpotQA development set \citep{yang2018hotpotqa}, a widely recognized benchmark for multi-hop reasoning, to measure the component's general capability against an objective external standard. Second, we evaluate the module on MQUAKE-CF-3K, one of our primary task datasets, to confirm its effectiveness directly within our framework's context.
The results are summarized in Table \ref{tab:correlation_evaluator}. The module achieves an Acc of 85.63\% on HotpotQA and 76.97\% on MQUAKE-CF-3K. The strong performance across both a general benchmark and our task-specific dataset confirms that the correlation evaluator is robust and well-suited for its role within the MIRAGE framework.

\begin{table}[!t]
\centering
\small
\captionsetup{width=\columnwidth}
\setlength{\tabcolsep}{4pt}
\caption{Acc (\%) of the Correlation Evaluator on a standard benchmark and a task-specific dataset.}
\label{tab:correlation_evaluator}
\begin{tabular*}{\columnwidth}{@{\extracolsep{\fill}} l cc @{}}
\toprule
\textbf{Dataset} & \textbf{Total Count} & \textbf{Acc (\%)} \\
\midrule
HotpotQA     & 7405 & 85.63 \\
MQUAKE-CF-3K & 3000 & 76.97 \\
\bottomrule
\end{tabular*}
\end{table}

\subsection{Case Study: Reasoning Failures and Corrections}
\label{app:case_study}
To better understand the effectiveness of our proposed method, we conduct a qualitative analysis of the reasoning chains generated by the baseline and our model. Table \ref{tab:case_study} illustrates three typical failure modes of the baseline model when handling edited knowledge, and how our approach successfully corrects them.

\begin{table*}[htbp]
    \centering
    \caption{Qualitative comparison of reasoning chains between the baseline and our proposed method. The baseline exhibits three typical failure modes: overconfidence (skipping multi-hop reasoning), repetitive matching (falling into a reasoning loop), and missing sub-questions (failing to extract intermediate entities). Our method successfully decomposes the question and applies the edited knowledge (\newans{Europe}).}
    \vspace{0.5em}
    \renewcommand{\arraystretch}{1.3}
    
    \begin{tabularx}{\textwidth}{@{} l X @{}}
        \toprule
        \textbf{Input Question} & Which continent does Celine Dion's country of citizenship belong to? \\
        \midrule
        \textbf{Edit-hop} & \textbf{[Edit Question]} Which continent is Canada located in? \\
        & \textbf{[Edited Knowledge]} Canada is located in \newans{Europe} \textit{(Ground Truth: \oldans{North America})}. \\
        & \textbf{[Expected Final Answer]} \newans{Europe}\\
        \bottomrule
    \end{tabularx}

    \vspace{1em}

    \begin{tabularx}{\textwidth}{@{} X c @{}}
        \toprule
        \textbf{Case Study of Typical Flaws} & \textbf{Final Answer} \\
        \midrule
        \multicolumn{2}{@{}l}{\textbf{Case 1: Overconfidence (Shortcut Reasoning)}} \\
        \midrule
        \textbf{[Direct Answer]} The entity of Subquestion is: North America. $\rightarrow$ \errortext{Ignored edited knowledge.} & \xmark\ \oldans{North America} \\
        \midrule
        \multicolumn{2}{@{}l}{\textbf{Case 2: Repetitive Matching (Reasoning Loop)}} \\
        \midrule
        \textbf{[Sub-Q 1]} What is the country of citizenship of Celine Dion? $\rightarrow$ Canada. \newline \textbf{[Sub-Q 2]} Which continent is the country of citizenship for Celine Dion, Canada, located in? $\rightarrow$ \errortext{Canada. The country of citizenship of Celine Dion is Canada.} & \xmark\ \oldans{North America} \\
        \midrule
        \multicolumn{2}{@{}l}{\textbf{Case 3: Missing Sub-questions (Reasoning Break)}} \\
        \midrule
        \textbf{[Sub-Q 1]} What is the country of citizenship of Celine Dion? $\rightarrow$ \errortext{(Singer)} \newline \textbf{[Sub-Q 2]} What country is Celine Dion a citizen of? $\rightarrow$ \errortext{(Singer)} \newline $\rightarrow$ \errortext{The information provided is insufficient to determine.} & \xmark\ None \\
        \midrule
        \multicolumn{2}{@{}l}{\textbf{The Revised Case}} \\
        \midrule
        \textbf{[Sub-Q 1]} What is the country of citizenship of Celine Dion? $\rightarrow$ Canada. \newline \textbf{[Sub-Q 2]} Which continent is Canada located in? $\rightarrow$ \newans{Europe}. & \cmark\ \newans{Europe} \\
        \bottomrule
    \end{tabularx}

    \label{tab:case_study}
\end{table*}

Furthermore, to intuitively illustrate the vulnerability of LLMs when confronted with complex, multi-hop knowledge conflicts, the following subsection presents a real-world instance from the financial domain of our NEXUS dataset, demonstrating how the compounding cognitive load of maintaining multiple counterfactual premises can trigger a complete collapse of the reasoning chain---a ``Reasoning Break''---even when earlier hops are answered correctly.

\subsection{Additional Case Study: Reasoning Break in the Financial Domain}
\label{app:financial_case}

To intuitively illustrate the vulnerability of LLMs when confronted with
complex, multi-hop knowledge conflicts, Table
\ref{tab:real_case_conflict} presents a real-world instance from the financial
domain of our NEXUS dataset. In this case, we introduce two counterfactual
edits (Phase 1): altering the sector with the highest risk spillover from the
\textit{Securities Industry} to the \textit{Insurance Industry}, and shifting
the strongest risk-absorbing sector from the \textit{Banking Industry} to the
\textit{Securities Industry}. As shown in the reasoning trajectory (Phase 2),
the model successfully overrides its parametric memory to answer the first two
sub-questions based on the injected edits. However, when attempting the final
hop, the compounding cognitive load of maintaining multiple counterfactual
premises triggers a complete collapse of the reasoning chain. The model
exhibits a severe hallucination---abruptly generating entirely irrelevant text
about the head of state in Croatia---before defaulting to an incorrect final
answer. This qualitative evidence strongly validates that our dataset
construction pipeline can effectively expose the ``Reasoning Break'' flaw in
current LLMs.

\begin{table*}[!t]
\centering
\caption{A real-world example of Conflict Data construction and the resulting LLM reasoning failure (Reasoning Break) from the NEXUS dataset.}
\label{tab:real_case_conflict}
\renewcommand{\arraystretch}{1.4}
\begin{tabularx}{\textwidth}{@{}p{3.2cm}X@{}}
\toprule
\textbf{Original Question} & In the Chinese financial system, which type of financial institution is identified as having the highest risk spillover intensity, and which type primarily plays a risk-absorbing role? \\
\midrule
\multicolumn{2}{@{}p{\dimexpr\textwidth-2\tabcolsep\relax}@{}}{\textbf{Phase 1: Edit-hop (Counterfactual Modification)}} \\
\midrule
\textbf{Fact 1 (Spillover)} & Highest risk spillover: \textcolor{gray}{\textit{[Ground Truth: Securities Industry]}} $\rightarrow$ \textcolor{blue}{\textbf{[Edited: Insurance Industry]}} \\
\textbf{Fact 2 (Absorption)} & Strongest risk absorption: \textcolor{gray}{\textit{[Ground Truth: Banking Industry]}} $\rightarrow$ \textcolor{blue}{\textbf{[Edited: Securities Industry]}} \\
\midrule
\multicolumn{2}{@{}l}{\textbf{Phase 2: Multi-hop Reasoning Evaluation (LLM Trajectory)}} \\
\midrule
\textbf{Sub-question 1} & In the Chinese financial system, which type of financial institution exhibits the most obvious risk diffusion and the highest risk spillover intensity? \\
\textbf{LLM Output 1} & \textit{...the entity of Subquestion is:} \textcolor{blue}{\textbf{Insurance Industry}} \hfill \textcolor{green!70!black}{\textbf{\ding{51} Hop 1: Success}} \\
\midrule
\textbf{Sub-question 2} & In the Chinese financial system, which type of financial institution presents a "risk-absorbing" systemic importance and has the strongest risk absorption capacity? \\
\textbf{LLM Output 2} & \textit{...the entity of Subquestion is:} \textcolor{blue}{\textbf{Securities Industry}} \hfill \textcolor{green!70!black}{\textbf{\ding{51} Hop 2: Success}} \\
\midrule
\textbf{Sub-question 3} & In the Chinese financial system, what role does the banking industry primarily play in systemic risk? \\
\textbf{LLM Output 3} & \textcolor{red}{\textit{The head of state in Croatia is Zoran Milanović. Final answer: Risk diffusion type}} \hfill \textcolor{red}{\textbf{\ding{55} Hop 3: Break}} \\
\bottomrule
\end{tabularx}
\end{table*}

\section{Limitations and Broader Impact}
\label{app:limitations}

\subsection{Limitations}
Our work has several limitations that we would like to acknowledge.
\begin{itemize}
    \item \textbf{P(True) threshold sensitivity.} The P(True) routing relies on
    a fixed threshold (0.5 by default). While the empirical distributions of
    incremental and conflict data are well separated, a task-specific threshold
    may further improve performance; automatic threshold selection is left for
    future work.
    \item \textbf{Reliance on instruction following.} Both the CDR and the TGC
    correct errors through targeted prompts. Their effectiveness therefore
    depends on the base model's instruction-following ability, which can vary
    across model families and scales.
    \item \textbf{Scope of evaluation.} Experiments are conducted on four LLMs
    and two primary benchmarks. The NEXUS dataset is primarily English-centric
    and includes a baseline geographic subset built from fictional entities;
    the extent to which the findings transfer to other languages and domains
    requires further investigation.
    \item \textbf{Statistical significance.} We do not report error bars or
    significance tests in the main paper; adding repeated-run variance
    estimates is part of future work.
    \item \textbf{Trust in retrieved knowledge.} As a RAG-based framework,
    MIRAGE inherits the assumption that the retrieved documents are reliable.
    Malicious or incorrect retrieval content could be propagated into the
    reasoning chain.
\end{itemize}

\subsection{Broader Impact}
This work is foundational research aimed at improving the robustness and
reliability of multi-hop reasoning in dynamically evolving knowledge
environments. The NEXUS benchmark and the MIRAGE framework can benefit
applications that depend on up-to-date knowledge, such as financial news
monitoring, scientific literature tracking, and medical information retrieval,
by helping models correctly incorporate both updated and novel facts.

As with any retrieval-augmented system, a potential negative societal impact is
that incorrect or deliberately injected retrieval content could be trusted and
propagated by the model. We encourage practitioners to pair MIRAGE with
retrieval quality assurance mechanisms. We do not foresee a direct path from
this work to malicious applications beyond those already shared by general
question-answering systems.

\end{document}